\documentclass[runningheads]{llncs}
\usepackage{verbatim} 
\usepackage{graphicx}
\usepackage{amssymb}

\begin{document}
\title{Adaptive Image-Feature Learning for Disease Classification Using Inductive Graph Networks}

\titlerunning{Adaptive Image-Feature Learning Using Inductive Graph Networks}

\author{
Hendrik Burwinkel\inst{1}
\and
Anees Kazi\inst{1}
\and
Gerome Vivar\inst{2}
\and
Shadi Albarqouni\inst{1}
\and
Guillaume Zahnd\inst{1}
\and
Nassir Navab\inst{1,3}
\and
Seyed-Ahmad Ahmadi\inst{2}
}
%
\authorrunning{H. Burwinkel et al.} 
\institute{
Computer Aided Medical Procedures, Technische Universit{\"a}t M{\"u}nchen, Boltzmannstra{\ss}e 3, 85748 Garching bei M{\"u}nchen, Germany\\
\email{hendrik.burwinkel@tum.de}
\and
German Center for Vertigo and Balance Disorders, Ludwig-Maximilians Universit\"at M\"unchen, Marchioninistr. 15, 81377 M\"unchen, Germany
\and
Computer Aided Medical Procedures, Johns Hopkins University, 3400 North Charles Street, Baltimore, MD 21218, USA
}

\maketitle              
\begin{abstract}
Recently, Geometric Deep Learning (GDL) has been introduced as a novel and versatile framework for computer-aided disease classification. GDL uses patient meta-information such as age and gender to model patient cohort relations in a graph structure. Concepts from graph signal processing are leveraged to learn the optimal mapping of multi-modal features, e.g. from images to disease classes. Related studies so far have considered image features that are extracted in a pre-processing step. We hypothesize that such an approach prevents the network from optimizing feature representations towards achieving the best performance in the graph network. We propose a new network architecture that exploits an inductive end-to-end learning approach for disease classification, where filters from both the CNN and the graph are trained jointly. We validate this architecture against state-of-the-art inductive graph networks and demonstrate significantly improved classification scores on a modified MNIST toy dataset, as well as comparable classification results with higher stability on a chest X-ray image dataset. Additionally, we explain how the structural information of the graph affects both the image filters and the feature learning.

\end{abstract}
\section{Introduction}
In the past few years, the potential of data processing with Geometric Deep Learning (GDL) on non-Euclidian data structures has been demonstrated by a large body of work~\cite{Bronstein2017,Defferrard2016,Kipf2016,Monti2017a,Parisot2017}. One uprising application, which is also the subject of this work, lies in the field of computer-aided disease classification (CAD), where the usage of patient meta-information is leveraged to build a graph system of meaningful patient inter-relations~\cite{Parisot2017,Kazi2018a}. 
These patient relations allow e.g. an improved classification of image data, which enables the prediction of a disease.
This is achieved by using various forms of graph signal processing for the training of localized graph filters, where patients are modeled as vertices in the graph, and images as feature vectors. One family of approaches \cite{Bronstein2017,Kipf2016,Parisot2017} are graph spectral methods, which utilize the normalized graph Laplacian $L$ to perform a Fourier transform in order to find optimal filters in the frequency domain. A major drawback of spectral methods is the limitation of their learned representation to the graph they are trained on. Therefore, their transductive approach requires a retraining for every new graph system, making them difficult to use for a direct application on newly received patient data. We therefore focus on non-spectral methods, which overcome this limitation by working on local graph structures, mimicking the standard convolution operation on the spatial relations of a graph node. The concept relevant for this work are composition-based spatial methods that use multiple consecutive layers to update the vertex representation. Hamilton et al. presented GraphSage \cite{Hamilton2017}, which samples a fixed number of neighbors of a vertex and aggregates them to receive an updated representation of the feature vector in every layer. This inductive approach allows the successive processing of every vertex of the graph. The system however did not define a preference of which neighboring feature representations are most relevant for the update. We thus focus on and adapt Graph Attention Networks (GAT) in this work. GAT was introduced by \cite{Velickovic2018} and leverages an attention mechanism to learn this importance measure in order to optimize the representation learning.\\
Regarding CAD systems, in \cite{Parisot2017} a GDL-approach for disease classification was proposed, modeling patient similarities through meta-information and performing patient classification in the ABIDE and ADNI datasets using Graph Convolutional Networks (GCN). In 2018, self-attention was used in \cite{Kazi2018a} to optimize the usage of meta-information in the TADPOLE dataset. Both methodologies used pre-learned features in their approach. Since we are evaluating our method on chest X-ray images, we are also mentioning related works out of this field. In 2017, Wang et al. \cite{Wang} proposed a new chest X-ray database, known as ``ChestX-ray8'', which contains 108,948 frontal view X-ray images of 32,717 unique patients for eight different disease classifications and performed multi-label disease classification and localisation using class activation mapping \cite{Zhou2015}. In \cite{Islam2017}, disease classification was realized on three chest X-ray datasets using standard deep convolutional networks, localizing activations with occlusion mapping. CheXNet \cite{Rajpurkar2017}, a DenseNet of 121-layers, was used for classification of all 14 diseases for the NIH ChestX-ray 14 dataset, with an emphasis on pneumonia. All the works relied on standard CNN approaches and did not leverage any non-imaging information.\\
\textbf{Contributions.} Although GDL approaches used for CAD showed promising performance, two major drawbacks are present in the methodologies developed so far. First, it is necessary to extract the features in a pre-processing step, which results in additional effort. Second, a potentially not optimized feature representation for the usage inside the graph system is learned. To address these limitations we propose CNNGAT, a novel method that yields an inductive end-to-end approach for both learning an optimized feature representation and performing the graph convolutions during training time. Additionally, the introduced concept leverages inter-class connections to specifically improve the performance on instances from classes which are more diffcult to distinguish. To the best of our knowledge this is the first work that combines the training of the CNN and the graph network to obtain optimized feature representations. We evaluate our proposed method on the ChestX-ray 14 \cite{Rajpurkar2017} and a modified MNIST dataset.

\section{Methodology}
\subsection{Explanation of the Proposed Model}
\begin{figure}[t]
    \centering
    \includegraphics[trim = 0px 0px 0px 0px, width=1.0\textwidth]{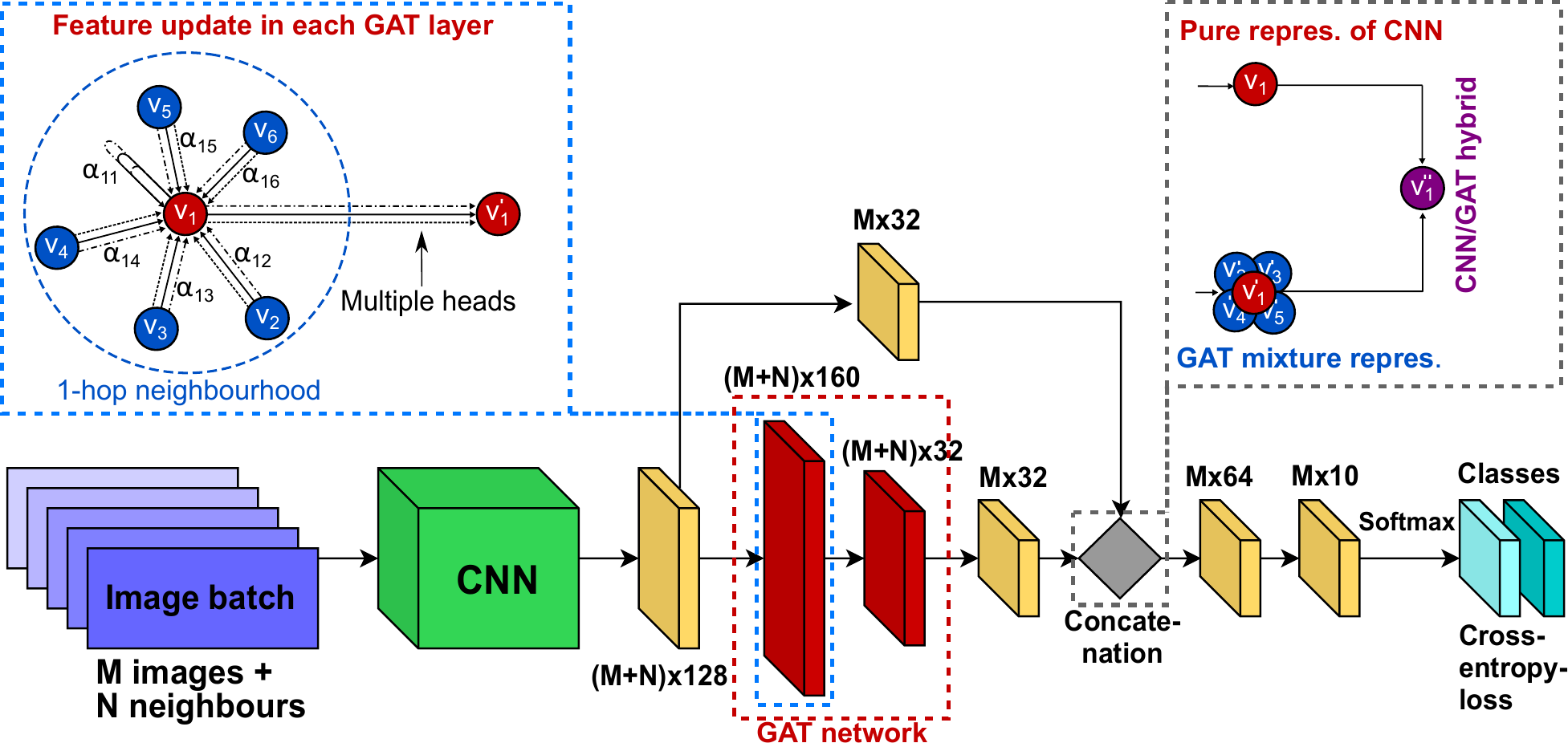}
    \caption{CNNGAT for a classification of 10 classes. An image batch of $M$ images and $N$ neighboring images is loaded and processed by the CNN. The extracted features are passed through the GAT layers. Then, only the M main batch representations resulting from the GAT network and CNN are concatenated and used to perform the final classification and loss backpropagation.}
    \label{fig:CNNGAT}
\end{figure}
\textbf{General Framework.} Our proposed network CNNGAT performs the classification of images represented as 2D intensity arrays $\textbf{X}$ using non-imaging meta-information in an inductive end-to-end approach. It tries to optimize the objective function $f(\textbf{X},G(\textbf{V},\textbf{E})): \textbf{X} \rightarrow \textbf{Y}$, where $G(\textbf{V}, \textbf{E})$ is a graph with vertices $\textbf{V}$ containing a feature representation of every image $\vec{x}_i$ and binary edges $\textbf{E}$ as connections between the vertices and $\textbf{Y}$ as set of classes. The whole set of feature representations for every vertex $\vec{v}_i$ of the graph is therefore defined as: $\textbf{V} = \{\vec{v}_1, \vec{v}_2,...,\vec{v}_N\}, \vec{v}_i \in \mathbb{R}^{F}$, where $F$ is the dimension of the representation. The edges $\textbf{E}$ are created based on non-imaging information belonging to every image $\vec{x}_i$. The CNN extracts a feature representation of image $\vec{x}_i$ and distributes it to the corresponding vertex $\vec{v}_i$. Subsequently, the obtained representation is processed by two consecutive GAT layers. The resulting representation is concatenated with the initial one of the CNN to perform the final prediction (Fig.~\ref{fig:CNNGAT}).\\
\textbf{Neighborhood Concept.} A GAT layer updates the feature representation $\vec{v}_i$ based on the ones of neighboring vertices. Therefore, it requires that a neighborhood of images is processed together with $\vec{x}_i$. Since the general size of image datasets prohibits the use of all neighbors for a vertex $\vec{v}_i \in \textbf{V}$, different from \cite{Velickovic2018}, only a fixed sub-sample of size $n$ of all vertices $\vec{v}_j$ with $e_{ij} \in \textbf{E}$ is considered as neighborhood $N_i$ of $\vec{v}_i$. If for a vertex $\vec{v}_i$, the size $n$ of the neighborhood cannot be reached, a random selection of $N_i$ is used multiple times until the amount $n$ of vertices is achieved. As described in \cite{Hamilton2017}, this procedure has the advantage of setting the computational effort of the network to a fixed amount.\\
\textbf{GAT Layer.} To achieve a higher representation of $\vec{v}_i \in \mathbb{R}^{F}$ with new dimension $F'$, a shared learnable linear transformation $\textbf{W} \in \mathbb{R}^{F' \times F}$ is applied to $\vec{v}_i$ and its considered neighbors $\vec{v}_j \in N_i$. Then, a shared attention mechanism $a$ that consists of a single-layer feed-forward network is used to obtain the attention coefficient $\alpha$, representing how important $\vec{v}_j$ is for the update of $\vec{v}_i$ and computed as $a(\textbf{W} \vec{v}_i, \textbf{W} \vec{v}_j) = \vec{a}^T [\textbf{W} \vec{v}_i|| \textbf{W} \vec{v}_j]$, with the concatenation (symbolized as $[ ~|| ~]$) of $\textbf{W} \vec{v}_i$ and $\textbf{W} \vec{v}_j$ and  $\vec{a} \in \mathbb{R}^{2F'}$. After activating the calculated attention using the activation $\sigma =$ LeakyReLU, the softmax function is applied for all $\vec{v}_j \in N_i$ to obtain normalized and easily comparable attention coefficients $\alpha$:
\begin{equation}
\alpha_{ij} = \frac{\exp (\sigma(\vec{a}^T([\textbf{W} \vec{v}_i|| \textbf{W} \vec{v}_j])))}{\sum_{r \in N_i} \exp (\sigma(\vec{a}^T [\textbf{W} \vec{v}_i|| \textbf{W} \vec{v}_r]))}
\end{equation}
Every attention coefficient is multiplied with the feature representation $\textbf{W} \vec{v}_j$ and added up to obtain the new representation $\vec{v'}_i$. To stabilize the representation, this procedure is repeated multiple times with individual $\textbf{W}^k$, called heads, performing the same attention mechanism for a vertex $\vec{v}_i$. The representations $\vec{v'}_i$ are concatenated (represented as $\Vert$), yielding the new feature representation:
\begin{equation}
    \vec{v'}_i = \Vert_{k=1}^K \sigma \left( \sum_{j \in N_i} \alpha_{ij}^k \textbf{W}^k \vec{v}_j \right)~,
\end{equation}
where $\alpha_{ij}^k$ is the attention coefficient of head $k$ for the vertices $\vec{v}_i$ and $\vec{v}_j$, and $K$ is the number of heads \cite{Velickovic2018}.\\ 
\textbf{CNN/GAT Hybrid.} The initial representation $\vec{v}_i$ is processed at the same time in a skip connection using a single-layer feed-forward network $\textbf{W}_{skip}$. The output of the layer is activated using LeakyReLU and then concatenated with the output of the graph attention network $\vec{v'}_i$ to receive the hybrid representation $\vec{v''}_i = [\vec{v}_i||\vec{v'}_i]$. This strategy enables both the initial representation of the CNN and the altered representation after neighborhood interaction to be used for the prediction, leveraging both information. After a final activation, a last single-layer feed-forward network is used to map the concatenated vector to the class output and apply a softmax function. The gradients are backpropagated through the whole network, updating not only the graph network but also the CNN, and therefore updating the feature extraction in every iteration.

\subsection{Motivation of Skip Connection for CNN/GAT Hybrid}
Two concepts motivate the concatenation of the representations extracted from the CNN and the GAT. First, the skip connection allows a direct gradient propagation to the CNN, thus fortifying a proper filter learning. More importantly, the combination of both outputs enables a comparison of the pure feature extraction of the CNN and the ``impure'' feature representation after its interaction with the neighborhood (Fig.~\ref{fig:CNNGAT} top right). The individual connectivity of instances from a class to other classes in the graph can yield an additional unique contribution to the feature representation after the feature aggregation process of GAT, if these inter-class connections are different for every class. Especially for classes usually difficult to distinguish, this individual contribution of feature representations from other instances significantly improves the classification performance on these classes since their originally similar feature representations are altered and more distinct compared to their initial representation transported by the skip-connection. Therefore, the interesting and unintuitive observation is obtained that inter-class connections can be highly beneficiary in the used setup.

\section{Experiments}
\subsection{Datasets}
First, a modified version of the MNIST dataset is used to prove the concept of the proposed method. MNIST consists of 70,000 handwritten digits from $0$ to $9$. Only the lower half of every image is used as image input, while the upper half is processed for the creation of the affinity graph. The objective is to show the improved classification of classes difficult to distinguish like \texttt{3} and \texttt{5} due to their identical lower shape. Following the inter-class setup explained in Sec. 2.2, we additionally take into account the digit \texttt{6}, which shows a similarity to only the number \texttt{5} in the upper half. We therefore create a subset of 6,000 images of the numbers \texttt{3}, \texttt{5} and \texttt{6} for training and 2,860 images for testing.\\
Secondly, we adapt the approach to the task of disease classification on the NIH ChestX-ray14 dataset. The dataset consists of 112,120 labelled X-ray images, containing one or more of 14 different diseases, with substantial imbalance for some classes. 16,000 single-label images of the eight diseases suggested by \cite{Wang} are taken into consideration to obtain a potentially clearer patient cohort system.\\

\subsection{Experimental Setup}
\textbf{Affinity Graph Construction.} For MNIST, the construction of the affinity graph is based on the upper half of the image. As a metric, the $l1$-distance of the flattened image intensity vectors is used. If the distance between the two image vectors $v_1$ and $v_2$ is below the threshold $\theta$, the edge is established. The best threshold was found with $\theta = 0.1$ for the average pixel intensity difference on an intensity scale from 0 to 1. Then, to analyze the proposed network and to prove that the obtained improvement is related to the previously described meaningful inter-class connections, three additional settings are used: 1)~random edges, 2)~edges based on label information, and 3)~edges based on labels including the helpful inter-class connection between \texttt{5} and \texttt{6}. It has to be stated that the usage of label information is just used here to prove the concept of the network.\\
For the NIH chest X-ray dataset, the accessible meta-information is used to create the affinity graph. Here, we are following the current state of the art introduced by \cite{Parisot2017} to leverage an age and gender difference metric. After performing an analysis of the statistics of the age and gender distribution with respect to the patients' diseases, we connect subjects in our experiments if at least one of the following three conditions is fulfilled for two patients $\textsc{p}_1$ and $\textsc{p}_2$:\\
1) patient id$_{\textsc{p}_1}$ = patient id$_{\textsc{p}_2}$\\
2) gender$_{\textsc{p}_1}$ = gender$_{\textsc{p}_2}$ \texttt{and} $|$age$_{\textsc{p}_1}$ - age$_{\textsc{p}_2}| \leq 1y$\\
3) gender$_{\textsc{p}_1}$ $\neq$ gender$_{\textsc{p}_2}$ \texttt{and} age$_{\textsc{p}_1}$ = age$_{\textsc{p}_2}$\\
\textbf{Network Setup.} We create three baselines and one alternative method to analyze the different aspects of CNNGAT. To show the improved learning of feature representations we compare against the performance of a simple CNN, the  normal GAT network on pre-learned features by the CNN (RawGAT), and our CNNGAT with pre-learnt features instead of end-to-end training (SkipGAT). To validate the hybrid approach we compare against our end-to-end CNNGAT network without using skip connections (EndGAT).\\
\textbf{Feature Extraction.} For MNIST, the CNNGAT was trained with a simple CNN containing two convolutional layers (32 and 64 channels). For the NIH dataset, AlexNet was used to easily prevent overfitting on the 16,000 images.\\
\textbf{Parameters.} The following parameters were chosen:\\
\underline{MNIST}: 60 features as GAT input, 2 GAT layers (30 and 10 units), 5 heads, dropout: 0.3, neighbors: 4, weight decay: 5e-3, lr: 0.02, lr $\times$ 0.3 every 20 epochs.\\
\underline{NIH}: 60 features as GAT input, 2 GAT layers (30 and 32 units), 5 heads, dropout: 0.3, neighbors: 4, weight decay: 5e-3, lr: 0.01, lr $\times$ 0.3 every 30 epochs.

\subsection{Modified MNIST Dataset}

\begin{figure}[t] 
\centering
\includegraphics[trim = 0px 50px 0px 60px, width=0.98\textwidth]{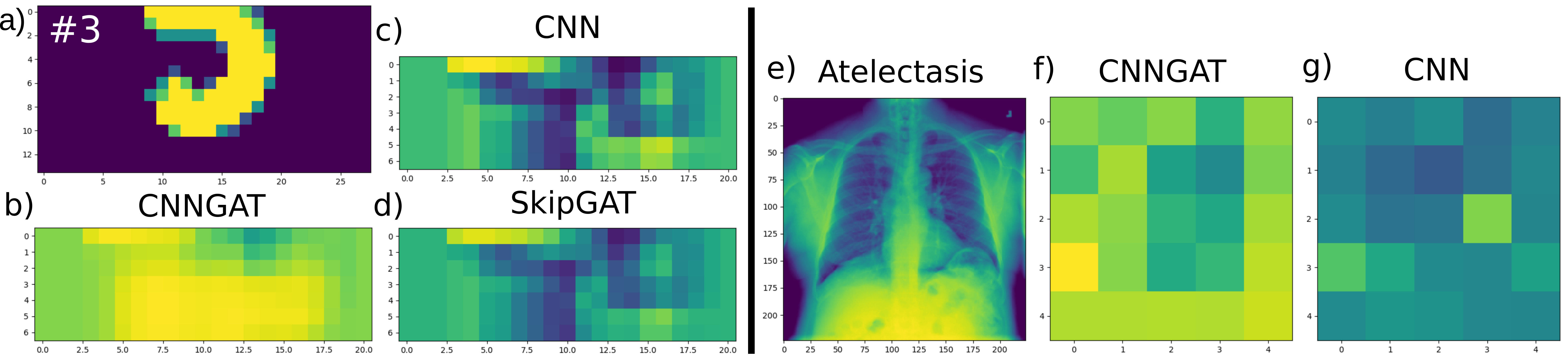}
\caption{
\textbf{Left}:
(a)~MNIST number \texttt{3}.
(b--d)~Occlusion shift on (a) with $7 \times 7$ window and stride length (SL) 1. Color bar corresponds to probability in softmax function for correct class when occ. is performed in that image region, the brighter the higher the probability (bright yellow: highest achieved probability for compared occlusions).
\textbf{Right}:
(e)~Atelectasis.
(f--g)~Occ. shift on (e) with $50 \times 50$ window and SL 50.}
\label{fig:Occ_shift}
\end{figure}

The results on the above described MNIST toy dataset shown in Tab.~\ref{tab:MNIST_res} indicate that the CNNGAT is indeed improving the classification by using the inter-class connections for numbers \texttt{5} and \texttt{6}. Using pre-trained features in the proposed network approach (SkipGAT) diminishes the performance of the system, showing the value of adaptive feature learning. Also the approach without Skip-Connection (EndGAT) is significantly outperformed (Wilcoxon signed-rank test, significance level $p<0.05$). To give further evidence that the features are learned in a more robust way, we perform a shifted occlusion for 1000 random images. A occluding window is slid across the image and after prediction is performed, the probability of the correct class in the softmax function is recorded for every position. One example is shown in Fig.~\ref{fig:Occ_shift} a-d) for an instance of the number \texttt{3}. It is visible that the classification is significantly more stable compared to the other networks, also stated in Tab.~\ref{tab:MNIST_res}.\\
For the second setting of graph edges described in Sec. 3.2, the lower part of Tab.~\ref{tab:MNIST_res} clearly shows that the presence of meaningful inter-class connections is leading to top-level performance even outperforming a clean label graph.

\begin{table}[t]
\begin{center}
\caption{Performance on MNIST dataset. Aff. shows which affinity mechanism was used: $\theta$= l1-distance, L=labels, CL=labels with connection between \texttt{5} and \texttt{6}. P-val is reported against CNNGAT. Occ. shows average correct class probability of softmax function for occlusion shift. O. p-val shows p-value compared to CNNGAT.}
\begin{tabular}{ |p{2cm}||p{1.0cm}|p{2.4cm}|p{2.4cm}|p{1.2cm}|p{1.0cm}|p{1.2cm}|  }
\hline
 Network & Aff. & Accuracy & Lowest class acc & p-val & Occ. & O. p-val\\
 \hline
 CNN   & / & 0.822 $\pm$ 0.005  & 0.69 $\pm$ 0.049 & 10e-5 & 0.69 & 0.0 \\
 RawGAT & $\theta$ & 0.794 $\pm$ 0.003 & 0.450 $\pm$ 0.008 & 10e-5 &  / &  / \\
 SkipGAT & $\theta$ & 0.829 $\pm$ 0.014 & 0.538 $\pm$ 0.066 & 10e-5 & 0.68 & 0.0 \\
 EndGAT & $\theta$ & 0.881 $\pm$ 0.016 & 0.780 $\pm$ 0.087 & 10e-4 & / & / \\
 \textbf{CNNGAT} & \textbf{$\theta$} & \textbf{0.911 $\pm$ 0.005} & \textbf{0.810 $\pm$ 0.035} & /  & \textbf{0.81} & / \\
 \hline
 CNNGAT & Rand. & 0.816 $\pm$ 0.004 & 0.613 $\pm$ 0.052 & /  & & \\
 CNNGAT & L & 0.980 $\pm$ 0.036 & 0.956 $\pm$ 0.079 & /  & & \\
 CNNGAT & \textbf{CL} & \textbf{0.992 $\pm$ 0.001} & \textbf{0.989 $\pm$ 0.003} & /  & & \\
 \hline
\end{tabular}
\label{tab:MNIST_res}
\end{center}
\end{table}

\subsection{NIH ChestX-ray 14 Dataset}
\begin{table}[b]
\begin{center}
\caption{Performance of CNN and CNNGAT on the described NIH ChestX-ray 14 dataset for 16,000 images. The column setting is identical to Tab. \ref{tab:MNIST_res}.}
\begin{tabular}{ |p{2cm}||p{1.0cm}|p{2.4cm}|p{2.4cm}|p{1.2cm}|p{1.0cm}|p{1.2cm}|  }
 \hline
 Network & Aff. & Accuracy & Lowest class acc & p-val & Occ. & O. p-val\\
 \hline
 CNN & / & 0.429 $\pm$ 0.015 & 0.144 $\pm$ 0.102 & 0.109 & 0.27 & 2.6e-5 \\
 \textbf{CNNGAT} & meta & \textbf{0.437 $\pm$ 0.014} & 0.086 $\pm$ 0.070 & / & \textbf{0.30} & / \\
 \hline
\end{tabular}
\label{tab:Chest_res}
\end{center}
\end{table}
Tab. \ref{tab:Chest_res} shows the results on the ChestX-ray 14 dataset. The CNNGAT had a slightly better performance than the used CNN, but the improvement is not significant. The absence of the effects seen on the MNIST dataset are addressed in the discussion. The occlusion shift analysis however shows that the classification is stabilized by the graph (Fig. \ref{fig:Occ_shift} right). The probability of predicting the correct class under occlusion is significantly higher compared to the raw CNN (Tab. \ref{tab:Chest_res}).

\section{Discussion and Conclusion}
The results on the MNIST dataset follow the expectation of the network performance. Only when training the CNN and GAT end-to-end, optimized performance is reached. Especially the comparison to SkipGAT, which is identical to CNNGAT except its pre-learned features, states the importance of the feature learning during training time. Additionally, the fact that the EndGAT setting is significantly outperformed fortifies the usage of the CNN/GAT hybrid approach. This behaviour was not clearly reproducible on the ChestX-ray 14 dataset. After performing a statistical analysis on the data distributions we hypothesize that the available meta-information was not relevant enough to build the required patient cohorts. Generating an artificial meaningful graph on the dataset showed the expected clear performance improvement.\\
We have proposed a new methodology to train a CNN and GAT network end-to-end and used their output in a hybrid approach to leverage both advanced feature learning and inter-class feature representations. The experiments clearly showed the superiority of the approach under the constraint that a meaningful graph can be created. In future work, the network architecture can be further optimized by e.g. including a learning mechanism for the adjacency system. With an automatized graph creation the necessity of finding meaningful meta-information is erased what makes the network applicable in a more general way.\\
\\
\textbf{Acknowledgements.} The study was supported by the Carl Zeiss Meditec AG in Oberkochen, Germany, and the German Federal Ministry of Education and Research (BMBF) in connection with the foundation of the German Center for Vertigo and Balance Disorders (DSGZ) (grant number 01 EO 0901). Further, we thank NVIDIA Corporation for the sponsoring of a Titan V GPU.

%
%
%
%
\bibliography{CNNGAT2019}

\end{document}